\documentstyle{nemlap}
\begin{document}
\title{
Learning Translation Rules From A Bilingual Corpus\footnote{This 
research has been supported in part 
by NATO Science for Stability Program Grant TU-LANGUAGE.}}

\author{Ilyas Cicekli and H. Altay G\"{u}venir} \
\institute{Dept. of Comp. Eng. and Info. Sc., Bilkent University \\
06533 Bilkent, Ankara, Turkey \\
{\tt e-mail: \{ilyas,guvenir\}@cs.bilkent.edu.tr}}

\maketitle

%
%

\begin{abstract}
This paper proposes a mechanism for learning
pattern correspondences between two languages from a corpus of
translated sentence pairs. The proposed mechanism uses analogical
reasoning between two translations. Given a pair of translations, the
similar parts of the sentences in the source language must correspond
the similar parts of the sentences in the target language. Similarly,
the different parts should correspond to the respective parts in the
translated sentences. The correspondences between the similarities,
and also differences are learned in the form of translation rules.
The system is tested on a small training dataset and produced 
promising results for further investigation.
\end{abstract}

%
%

\section{Introduction}
\label{introduction:sec}
Traditional approaches to machine translation (MT) suffer from
tractability, scalability and performance problems due to the
necessary extensive knowledge of both the source and the target
languages. Corpus-based machine translation is one of the alternative
directions that have been proposed to overcome the difficulties of
traditional systems. Two fundamental approaches in corpus-based MT have
been followed. These are {\em statistical} and {\em example-based} machine
translation (EBMT), also called {\em memory-based} machine translation
(MBMT). Both approaches assume the existence of a bilingual text (an
already translated corpus) to derive a translation for an input. While
statistical MT techniques use statistical metrics to choose the most probable
words in the target language, EBMT techniques employ pattern matching
techniques to translate subparts of the given input \cite{arnold-et-al}.

Exemplar-based representation has been widely used in 
Machine Learning (ML). 
According to Medin and Schaffer
\cite{medin-shaffer}, who originally proposed exemplar-based learning
as a model of human learning, examples are stored in memory without
any change in the representation. Here, an exemplar is a
characteristic example stored in the memory. The basic idea in
exemplar-based learning is to use past experiences or cases to
understand, plan, or learn from novel situations \cite{hammond,kolodner,ram}. 

EBMT has been proposed by Nagao \cite{nagao} as {\it Translation by
Analogy} which is in parallel with memory based reasoning
\cite{stanfill}, \mbox{case-based} reasoning \cite{reisbech} and
derivational analogy \cite{carbonell}. 
Example-based translation relies on the use of past translation
examples to derive a translation for a given input
\cite{furuse-iida,nirenburg,sato,sato-nagao,sumita-iida}. The input
sentence to be translated is compared with the example translations
analogically to retrieve the {\it closest} examples to the
input. Then, the fragments of the retrieved examples are translated 
and recombined in the target language. Prior to the translation of an
input sentence, the correspondences between the source and target
languages should be available to the system; however this issue has
not been given enough consideration by the current EBMT
systems. Kitano \cite{kitano} has adopted the manual encoding of the translation
rules, however this is a difficult and an error-prone task for a large
corpus. 
Wu \cite{wu} uses a method to extract phrasal translation
examples in sentence-aligned parallel corpora using
a probabilistic translation lexicon for the language pair.
Wu's {\em inversion transduction grammar} (ITG) formalism
is used to model two languages simultaneously.
In this paper, we formulate this acquisition
problem as a machine learning task in order to automate the process. 

In this paper, we propose a technique which stores exemplars in the
form of {\it templates} that are {\it generalized exemplars}.
A template is an example translation pair where some components (e.g.,
word stems and morphemes) are generalized by replacing them with
variables in both sentences, and establishing bindings between the
variables. We will refer this technique as GEBMT for {\it Generalized
Example Based Machine Translation}.
We assume no grammatical knowledge about languages
except morphological structure of some words in the languages.

The algorithm we propose here, for learning such templates, is based on
a heuristic to learn the correspondences between the patterns in the
source and target languages, from two translation pairs. The heuristic
can be summarized as follows: 
Given two translation pairs, if the sentences in the source language
exhibit some similarities, then the corresponding sentences in the
target language must have similar parts, and they must be translations
of the similar parts of the sentences in the source language.
Further, the remaining different parts of the source sentences should also
match the corresponding differences of the target sentences. However,
if the sentences do not exhibit any similarity, then no
correspondences are inferred. Consider the following translation pair
given in English and Turkish to illustrate the heuristic:
{\tt
\begin{tabbing}
\hspace*{5mm} \= \hspace*{10mm} \= \hspace*{5mm} \= \hspace*{10mm} \kill
\> \underline{I give+PAST the} book \underline{to Mary} \\
\>\>  $\leftrightarrow$ 
  \> \underline{Mary+DAT} kitap\underline{+ACC ver+PAST+1SG}
\\
\> \underline{I give+PAST the} pencil \underline{to Mary} \\
\>\> $\leftrightarrow$ 
  \> \underline{Mary+DAT} kur\c{s}un kalem\underline{+ACC ver+PAST+1SG}
\end{tabbing}
}

Similarities between the translation examples are shown as underlined.
The remaining parts are the differences between the sentences. We
represent the similarities in the source language as {\tt "I give+PAST the
$X^S$ to Mary"}, and the corresponding similarities in the
target language as {\tt "Mary+DAT $X^T$+ACC ver+PAST+1SG"}. 
According to our heuristic, these similarities should correspond each
other. Here, $X^S$ denotes a component that can be replaced by
{\it any} appropriate structure in the source language and $X^T$
refers to its translation in the target language. This notation
represents an {\it abstraction} of the differences {\{\tt book}
vs. {\tt pencil\}} and {\{\tt kitap} vs. {\tt kur\c{s}un kalem\}} in
the source and target languages, respectively. Using the heuristic
further, we infer that {\tt book} should correspond to {\tt kitap}
and {\tt pencil} should correspond to {\tt kur\c{s}un kalem}; hence
learning further correspondences between the examples. 

Our learning algorithm based on this heuristic is called TRL ({\it
Translation Rule Learner}). Given a corpus of translation cases,
TRL infers the correspondences between the source and target languages
in the form of translation rules. These rules can be used for translation
in both directions. Therefore, in the rest of the paper we will refer
these languages as $L_1$ and $L_2$. Although the examples and
experiments herein are on English and Turkish, we believe that the model is
equally applicable to other language pairs.

The rest of the paper is organized as
follows. Section\,\ref{learning:sec} describes the underlying
mechanisms of TRL, along with sample rule
derivations. 
Section\,\ref{examples:sec} gives more learning examples.
Section\,\ref{translation:sec} illustrates the 
translation process using translation rules. Section\,\ref{conclusion:sec}
concludes the paper.

\section{Learning}
\label{learning:sec}

Our learning algorithm TRL infers translation rules using 
similarities and differences between a pair of 
translation examples $ (E_i,E_j) $ from a bilingual corpus.
A translation example $ E $ is also a pair
$ (E^{L_1} \leftrightarrow E^{L_2}) $ where
$ E^{L_1} $ and $ E^{L_2} $ are equivalent sentences 
in languages ${L_1}$ and ${L_2}$.
Using a matching algorithm, we find 
a match sequence $ M^{L_1} $ representing similarities and
differences in $ E_i^{L_1} $ and $ E_j^{L_1} $, 
a match sequence $ M^{L_2} $ for $ E_i^{L_2} $ and $ E_j^{L_2} $.
From these two match sequences, we learn translation rules.

In our examples, we will use translation
examples between English and Turkish.
A translation example consists of an English
sentence and a Turkish sentence.
We will use the lexical level 
representation\footnote {In our examples, 
{\tt PAST, AOR, PRG, FUT} denote past, 
aorist, progressive and future tenses,
{\tt COND, NEC} denote necessitative and conditional,
{\tt ACC, DAT, LOC, ABL} denote accusative, dative, locative and 
ablative case markers for
nouns, {\tt 1SG, 2SG, 3SG} denote first, second and third singular
verbal agreements, {\tt COP} denotes copula in verbs. }
for each sentence in our examples.
For example, the English sentence 
{\em ``I broke a pencil''} will be represented by 
\begin{tabbing}
\hspace*{5mm} \= \hspace*{30mm} \= \hspace*{5mm} \= \hspace*{10mm} \kill 
\> {\tt I break+PAST a pencil}
\end{tabbing}
and its equivalent Turkish sentence
{\em ``Bir kur\c{s}un kalem k{\i}rd{\i}m''}
will be represented by 
\begin{tabbing}
\hspace*{5mm} \= \hspace*{30mm} \= \hspace*{5mm} \= \hspace*{10mm} \kill 
\> {\tt Bir kur\c{s}un kalem k{\i}r+PAST+1SG}
\end{tabbing}

For a pair of translation example
$ ((E_1^{L_1} \leftrightarrow E_1^{L_2}),(E_2^{L_1} \leftrightarrow E_2^{L_2})) $, 
the matching algorithm produces match sequences
$ M^{L_1} $ and $ M^{L_2} $ to represent
similarities and differences in examples in languages
$ {L_1} $ and $ {L_2} $, respectively.
A match sequence $ M $ for two different sentences
will be in the following form.
\begin{tabbing}
\hspace*{5mm} \= \hspace*{30mm} \= \hspace*{5mm} \= \hspace*{10mm} \kill 
\>  $ S_1\,D_1\,S_2\,\cdots\,D_n\,S_{n+1} $  where $ n \geq 1 $
\end{tabbing}
In that sequence, each $ S_i $ represents a similarity
between sentences.
In other words, it is a substring which is common in both of
those sentences.
Each $ D_i $ represents a difference which is 
a pair of non-empty substrings of sentences, one from
the first sentence and the other from the second sentence.
For each difference $D_i^1 : D_i^2$, $D_i^1$ and $D_i^2$ do not
contain any common item.
Also, no lexical item in a similarity $S_i$ appear in
any previously formed difference $D_k$ for $k<i$.
Any of $S_1$ or $S_{n+1}$ can be empty, 
however, $S_i$ for $1<i<n+1$ must be non-empty.
These restrictions guarantee tha there exists either
a unique match or no match between two different examples.

For example, in the following translation examples
\begin{tabbing}
\hspace*{5mm} \= \hspace*{30mm} \= \hspace*{5mm} \= \hspace*{10mm} \kill 
\> {\tt \underline{it is a} book} 	\> $\leftrightarrow$ 
  \> {\tt \underline{o bir} kitap\underline{+COP} } \\
\> {\tt \underline{it is a} pencil} 	\> $\leftrightarrow$ 
  \> {\tt \underline{o bir} kur\c{s}un kalem\underline{+COP} } 
\end{tabbing}
similarities are underlined and differences are not.
The match sequence for English sentences will be
\begin{tabbing}
\hspace*{5mm} \= \hspace*{30mm} \= \hspace*{5mm} \= \hspace*{10mm} \kill 
\>  {\tt \underline{it is a} book:pencil } 
\end{tabbing}
Note that we have one similarity and one difference between
English sentences.
The matching sequence for Turkish sentences will be
\begin{tabbing}
\hspace*{5mm} \= \hspace*{30mm} \= \hspace*{5mm} \= \hspace*{10mm} \kill 
\>  {\tt \underline{o bir} kitap:kur\c{s}un kalem \underline{+COP} }
\end{tabbing}
where we have two similarities and one difference.

In the example above, the difference in English sentences
must correspond to the difference in Turkish sentences,
and similarities in them must correspond to each other
in that context.
TRL can learn the following
translation rules from differences and similarities in
that example.
\begin{tabbing}
\hspace*{5mm} \= \hspace*{20mm} \= \hspace*{5mm} \= \hspace*{30mm} \= x \kill 
\> {\tt book}        \> $\leftrightarrow$ \> {\tt kitap}  \\
\> {\tt pencil}      \> $\leftrightarrow$ \> {\tt kur\c{s}un kalem } \\ 
\> {\tt it is} $X^E$ \> $\leftrightarrow$ \> {\tt o bir} $X^T$ {\tt +COP}
   \> where $X^E$ is a translation of $X^T$
\end{tabbing}
First two rules are learned from differences in English and 
Turkish sentences, namely
{\tt book:pencil} and {\tt kitap:kur\c{s}un kalem}.
The last rule is learned from similarities in the example.
In addition to these three learned rules, we also
put two translation rules directly given in the example
into our learned rule database.
Of course, they are more specific forms of the third 
learned rule.
We order rules from the most specific to the least specific
in the database.
During translation, the first applicable specific rule will
be used for the translation of a sentence
as a result of this ordering.

When the number of differences in two match sequences
$M^{L_1}$ and $M^{L_2}$ of a pair of translation examples
is greater than 1, say $n$, the learning algorithm
only learns new rules if $n-1$ differences can be resolved
using already learned rules from previous examples.
Otherwise, the current version of the algorithm
cannot learn new rules.
From the following example,
\begin{tabbing}
\hspace*{5mm} \= \hspace*{50mm} \= \hspace*{5mm} \= \hspace*{10mm} \kill 
\>  {\tt I \underline{give+PAST the} book} \> $\leftrightarrow$ 
 \> {\tt Kitap\underline{+ACC ver+PAST}+1SG } \\
\>  {\tt You \underline{give+PAST the} pencil} \> $\leftrightarrow$ 
 \> {\tt Kur\c{s}un kalem\underline{+ACC ver+PAST}+2SG } 
\end{tabbing}
we will get the following match sequences.
\begin{tabbing}
\hspace*{5mm} \= \hspace*{50mm} \= \hspace*{5mm} \= \hspace*{10mm} \kill 
\> $M^E$ = {\tt I:You \underline{give+PAST the} book:pencil}   \\
\> $M^T$ = {\tt Kitap:Kur\c{s}un kalem \underline{+ACC ver+PAST} +1SG:+2SG }
\end{tabbing}
Both $M^E$ and $M^T$ have two differences.
If we had not learned anything before this example,
there is no way to know whether the difference {\tt I:You}
in English sentences 
corresponds to the difference {\tt Kitap:Kur\c{s}un kalem} or
{\tt +1SG:+2SG} in Turkish sentences.
Now, let us assume that we have already learned
the following translation rules
from some previous examples.
\begin{tabbing}
\hspace*{5mm} \= \hspace*{15mm} \= \hspace*{5mm} \= \hspace*{10mm} \kill 
\>  {\tt book}   \> $\leftrightarrow$ \> {\tt Kitap} \\
\>  {\tt pencil} \> $\leftrightarrow$ \> {\tt Kur\c{s}un kalem} 
\end{tabbing}
Since we now know that the difference {\tt book:pencil} 
corresponds to the difference {\tt kitap:kur\c{s}un kalem},
the difference {\tt I:You} must correspond to 
the difference {\tt +1SG:+2SG}.
Thus, we can learn the following new translation rules
from this example. 
\begin{tabbing}
\hspace*{5mm} \= \hspace*{10mm} \= \hspace*{5mm} \= \hspace*{40mm} \= x \kill 
\> {\tt I}   \> $\leftrightarrow$ \> {\tt +1SG} 
   \> where $X^E$ is a translation of $X^T$  \\
\> {\tt You} \> $\leftrightarrow$ \> {\tt +2SG}  
   \> and $Y^E$ is a translation of $Y^T$. \\
\hspace*{5mm} \= \hspace*{42mm} \= \hspace*{5mm} \= \hspace*{20mm} \= x \kill 
\> $X^E$ {\tt give+PAST the} $Y^E$ \> $\leftrightarrow$ 
  \> $Y^T$ {\tt +ACC ver+PAST} $X^T$    
\end{tabbing}

For a given pair of translation examples,
$ ((E_1^{L_1} \leftrightarrow E_1^{L_2}),(E_2^{L_1} \leftrightarrow E_2^{L_2}))$,
the algorithm of the translation rule learner (TRL) for this pair 
is given in Figure~\ref{fig-learning-alg}.
In that algorithm, first we find match sequences $M^{L_1}$ and
$M^{L_2}$ for sentences in languages $L_1$ and $L_2$, respectively.
Then, we try to reduce the number of differences in
these match sequences to one.
At the same time, we construct $Condition$ which is a conjunction
of translation goals for a translation rule
which will be learned later in the algorithm.
After this reduction, each of our match sequences will have exactly
one difference.
So, these unlearned differences must correspond to each other.
From this fact, we learn three translation rules
given at the end of the algorithm.
In the implementation, each learned translation rule 
is represented in the form of a Prolog fact or rule.

\begin{figure}
\begin{tabbing}
\hspace*{5mm} \= \hspace*{2mm} \= \hspace*{2mm} \= \hspace*{2mm} \= x \kill 
Let $ ((E_1^{L_1} \leftrightarrow E_1^{L_2}),(E_2^{L_1} \leftrightarrow E_2^{L_2})$
be a pair of translation examples.
\\
$M^{L_1} \leftarrow match(E_1^{L_1},E_2^{L_1})$;
\\
$M^{L_2} \leftarrow match(E_1^{L_2},E_2^{L_2})$;
\\
{\bf if} $\#ofSimilarity(M^{L_1}) = 0$ {\bf or} $\#ofSimilarity(M^{L_2}) = 0 $
{\bf then} $exit$;
\\
{\bf if} $\#ofDifference(M^{L_1}) = 0$ {\bf or} 
\\ \>
$\#ofDifference(M^{L_1}) \neq \#ofDifference(M^{L_2}) $ {\bf then} $exit$;
\\
$Condition \leftarrow ``"$;
\\
$i \leftarrow 1$;
\\
{\bf while} $\#ofDifference(M^{L_1}) > 1$ {\bf do}
\\ \> 
{\bf begin}
\\ \>\> 
{\bf if} there exists a $D^{L_1}$ in $M^{L_1}$ and a $D^{L_2}$ in $M^{L_2}$ such that 
\\ \>\>\> 
the correspondence of $D^{L_1}$ to $D^{L_2}$ has been already learned {\bf then}
\\ \>\>\> 
{\bf begin}
\\ \>\>\>\> 
Replace $D^{L_1}$ in $M^{L_1}$ with a new variable $X_i^{L_1}$;
\\ \>\>\>\> 
Replace $D^{L_2}$ in $M^{L_2}$ with a new variable $X_i^{L_2}$;
\\ \>\>\>\> 
Add $``X_i^{L_1} \leftrightarrow X_i^{L_2} and"$ to the end of $Condition$;
\\ \>\>\>\> 
$i \leftarrow i + 1 $;
\\ \>\>\> 
{\bf end}
\\ \>\>
{\bf else} $exit$;
\\ \> 
{\bf end}
\\
Let $D^{L_1}$ in $M^{L_1}$ and $D^{L_2}$ in $M^{L_2}$ be unlearned differences such that
\\
$D^{L_1}$ is $D_1^{L_1}:D_2^{L_1}$ and $D^{L_2}$ is $D_1^{L_2}:D_2^{L_2}$;
\\
Replace $D^{L_1}$ in $M^{L_1}$ with a new variable $X_i^{L_1}$;
\\ 
Replace $D^{L_2}$ in $M^{L_2}$ with a new variable $X_i^{L_2}$;
\\ 
Add $``X_i^{L_1} \leftrightarrow X_i^{L_2}"$ to the end of $Condition$;
\\
Learn the following translation rules:
\\ \>
$D_1^{L_1} \leftrightarrow D_1^{L_2}$
\\ \>
$D_2^{L_1} \leftrightarrow D_2^{L_2}$
\\ \>
$M^{L_1} \leftrightarrow M^{L_2} \; if \; Condition$
\end{tabbing}
\caption{Translation Rule Learner Algorithm For Two Translated Sentence Pairs
\label{fig-learning-alg}}
\end{figure}

\section{Examples}
\label{examples:sec}
In order to evaluate the TRL algorithm we have developed a sample
bilingual parallel text. In this section, we will illustrate the
behavior of TRL on that sample text.

\vspace*{3mm}
\noindent{\bf Example 1:} Given the example translations 
``I saw you at the garden'' $\leftrightarrow$ 
``Seni bah\c{c}ede g\"{o}rd\"{u}m'' and 
``I saw you at the party'' $\leftrightarrow$
``Seni partide g\"{o}rd\"{u}m'', 
their lexical level representations are
{\tt
\begin{tabbing}
\hspace*{5mm} \= \hspace*{10mm} \= \hspace*{5mm} \= \hspace*{20mm} \= x \kill
\> \underline{i see+PAST you at the} garden $\leftrightarrow$ 
   \underline{sen+ACC} bah\c{c}e\underline{+LOC g\"{o}r+PAST+1SG}\\
\> \underline{i see+PAST you at the} party $\leftrightarrow$ 
   \underline{sen+ACC} parti\underline{+LOC g\"{o}r+PAST+1SG}\\
\end{tabbing}
}
\noindent
From these examples, the following translation rules are learned:
{\tt
\begin{tabbing}
\hspace*{5mm} \= \hspace*{10mm} \= \hspace*{5mm} \= \hspace*{20mm} \= x \kill
\> i see+PAST you at the $X_1^E$ $\leftrightarrow$ 
  sen+ACC $X_1^T$+LOC g\"{o}r+PAST+1SG 
\\ 
\>\> {\bf if} $X_1^E$  $\leftrightarrow$ $X_1^T$
\\
\> garden  $\leftrightarrow$  bah\c{c}e 
\\
\> party   $\leftrightarrow$  parti
\end{tabbing}
}

\vspace*{3mm}
\noindent{\bf Example 2:} Given the example translations 
``It is raining'' $\leftrightarrow$ ``Ya\u{g}mur ya\u{g}{\i}yor'', 
``He comes'' $\leftrightarrow$ ``Gelir'', 
``If it is raining then you should take an umbrella'' $\leftrightarrow$ 
``E\u{g}er ya\u{g}mur ya\u{g}{\i}yorsa bir \c{s}emsiye almal{\i}s{\i}n'' and 
``If he comes then we will go to the theater'' $\leftrightarrow$ 
``E\u{g}er gelirse tiyatroya gidece\u{g}iz'',
their lexical level representations are
{\tt
\begin{tabbing}
\hspace*{5mm} \= \hspace*{10mm} \= \hspace*{5mm} \= \hspace*{20mm} \= x \kill
\> it is rain+PRG $\leftrightarrow$ 
   ya\u{g}mur ya\u{g}{\i}+PRG  \\
\> He come+AOR $\leftrightarrow$ 
   gel+AOR \\ 
\> \underline{if} it is rain+PRG \underline{then} you should take an umbrella 
\\ \>\>   $\leftrightarrow$ 
   \underline{e\u{g}er} ya\u{g}mur ya\u{g}{\i}+PRG\underline{+COND} bir \c{s}emsiye al+NEC+2SG  \\
\> \underline{if} he come+AOR \underline{then} we will go to the theater
\\ \>\>$\leftrightarrow$ 
   \underline{e\u{g}er} gel+AOR\underline{+COND} tiyatro+DAT git+FUT+1PL
\end{tabbing}
}
\noindent
From the last two examples using first two examples,
the following translation rules are learned:
{\tt
\begin{tabbing}
\hspace*{5mm} \= \hspace*{10mm} \= \hspace*{5mm} \= \hspace*{20mm} \= x \kill
\> if $X_1^E$ then $X_2^E$ $\leftrightarrow$ e\u{g}er $X_1^T$+COND $X_2^T$ 
\\ \>\> {\bf if} $X_1^E \leftrightarrow X_1^T$ {\bf and}
   $X_2^E \leftrightarrow X_2^T$ \\
\> you should take an umbrella $\leftrightarrow$ bir \c{s}emsiye al+NEC+2SG  \\
\> we will go to the theater $\leftrightarrow$ tiyatro+DAT git+FUT+1PL
\end{tabbing}
}

\vspace*{3mm}
\noindent{\bf Example 3.} Given the example translations 
``I went'' $\leftrightarrow$ ``gittim'',
``you went'' $\leftrightarrow$ ``gittin'' and 
``I came'' $\leftrightarrow$ ``geldim'',
their lexical level representations are
{\tt
\begin{tabbing}
\hspace*{5mm} \= \hspace*{10mm} \= \hspace*{5mm} \= \hspace*{20mm} \= x \kill
\> i go+PAST $\leftrightarrow$ git+PAST+1SG  \\
\> you go+PAST $\leftrightarrow$ git+PAST+2SG  \\
\> i come+PAST $\leftrightarrow$ gel+PAST+1SG  
\end{tabbing}
}
\noindent
From the first and second examples where differences are 
{\tt i:you} and {\tt +1SG:+2SG}, 
the following translation rules are learned:
{\tt
\begin{tabbing}
\hspace*{5mm} \= \hspace*{10mm} \= \hspace*{5mm} \= \hspace*{20mm} \= x \kill
\> $X_1^E$ go+PAST $\leftrightarrow$ git+PAST $X_1^T$ 
\\ \>\> {\bf if} $X_1^E \leftrightarrow X_1^T$ \\
\> i $\leftrightarrow$ +1SG  \\
\> you $\leftrightarrow$ +2SG  \\
\end{tabbing}
}
\noindent
And from the first and third examples where differences are 
{\tt go:come} and {\tt git:gel}, 
the following translation rules are learned:
{\tt
\begin{tabbing}
\hspace*{5mm} \= \hspace*{10mm} \= \hspace*{5mm} \= \hspace*{20mm} \= x \kill
\> i $X_1^E$+PAST $\leftrightarrow$ $X_1^T$+PAST+1SG 
\\ \>\> {\bf if} $X_1^E \leftrightarrow X_1^T$ \\
\> go $\leftrightarrow$ git  \\
\> come $\leftrightarrow$ gel  \\
\end{tabbing}
}
\noindent

\section{Translation}
\label{translation:sec}
The translation rules learned by the TRL algorithm can be used in
the translation directly. The outline of the translation process is given
below:

\begin{enumerate}
\item 
First, the lexical level representation of the input sentence 
is derived. 
\item 
The most specific matching translation rule is found
for the input sentence.
If the template for the language of the input sentence
 in a translation
rule matches the input sentence, we call it a matching rule.
During this matching, certain variables in the template can
bind to substrings of the input sentence.
Then, translations for these bound variables are sought.
Thus, we will get the lexical level representation
of the output sentence if these processes are successful.
The most specific matching rule contains maximum number matching
terminals and minimum number of variables.
\item The surface level representation of the output sentence obtained in the
previous step is generated.
\end{enumerate}
 
Note that, if the input sentence in the source language is ambiguous, then
templates corresponding to each sense will be retrieved, and the
sentences for each sense will be generated. The translation rules
learned by TRL can be used for translation in both directions.

\section{Conclusion}
\label{conclusion:sec}
In this paper, we have presented a model for learning translation
rules between two languages. We integrated this model
with an example-based translation model into Generalized 
Exemplar-Based Machine Translation. 
We have implemented this model as the TRL
(Translation Rule Learner) algorithm. The TRL algorithm is
illustrated in learning translation rules between Turkish and
English. 

The major contribution of this paper is that the proposed TRL algorithm
eliminates  the need for manually encoding the translations, which is
a difficult task for a large corpus. The TRL algorithm can work
directly on surface level representation of sentences. However, in
order to generate useful translation patterns, it is helpful to use
the lexical level representations. It is usually trivial, at least for
English and Turkish, to obtain the lexical level representations of
words. 

Our main motivation was that the underlying inference mechanism is
compatible with one of the ways humans learn languages, i.e. learning
from examples. We believe that in everyday usage, humans learn general
sentence patterns, using the similarities and differences between many
different example sentences that they are exposed to. This observation
lead us to the idea that a computer can be trained similarly, using
analogy within a corpus of example translations. 

The accuracy of the translations learned by this approach is quite
high with ensured grammaticality. Given that a translation is carried
out using the rules learned, the accuracy of the output translation
critically depends on the accuracy of the rules learned. 

We do not require an extra operation to maintain the grammaticality
and the style of the output, as in Kitano's EBMT model \cite{kitano}. 
The information necessary to maintain these issues is directly 
provided by the translation rules. 

The model that we have proposed in this paper may be integrated with  
an intelligent tutoring system (ITS) for second language learning. 
The rule representation in our model provides a level of 
information that may help in error diagnosis and student modeling
tasks of an ITS. The model may also be used in tuning the teaching
strategy according to the needs of the student by analyzing the
student answers analogically with the closest cases in the
corpus. Specific corpora may be designed to concentrate on certain
topics that will help in student's acquisition of the target
language. The work presented by this paper provides an opportunity to
evaluate this possibility as a future work.

\end{document}